\documentclass[10pt,twocolumn,letterpaper]{article}

\usepackage{cvpr}             

\definecolor{cvprblue}{rgb}{0.21,0.49,0.74}
\usepackage[pagebackref,breaklinks,colorlinks,allcolors=cvprblue]{hyperref}

\usepackage[linesnumbered,ruled,vlined]{algorithm2e}
\usepackage{multirow}

\title{Mitigating Hallucinations in Diffusion Models\\through Adaptive Attention Modulation}

\author{Trevine Oorloff \qquad Yaser Yacoob \qquad Abhinav Shrivastava\\
University of Maryland,
College Park, Maryland, USA\\
{\tt\small \{trevine,yaser,abhinav2\}@umd.edu}
}

\begin{document}
\maketitle
\begin{abstract}
Diffusion models, while increasingly adept at generating realistic images, are notably hindered by hallucinations --- unrealistic or incorrect features inconsistent with the trained data distribution.    
In this work, we propose Adaptive Attention Modulation (AAM), a novel approach to mitigate hallucinations by analyzing and modulating the self-attention mechanism in diffusion models. 
We hypothesize that self-attention during early denoising steps may inadvertently amplify or suppress features, contributing to hallucinations. 
To counter this, AAM introduces a temperature scaling mechanism within the softmax operation of the self-attention layers, dynamically modulating the attention distribution during inference. 
Additionally, AAM employs a masked perturbation technique to disrupt early-stage noise that may otherwise propagate into later stages as hallucinations.
Extensive experiments demonstrate that AAM effectively reduces hallucinatory artifacts, enhancing both the fidelity and reliability of generated images. 
For instance, the proposed approach improves the FID score by 20.8\% and reduces the percentage of hallucinated images by 12.9\% (in absolute terms) on the Hands dataset. 
\end{abstract}    
\section{Introduction}
\label{sec:intro}

The ability of diffusion models \citep{ho2020ddpm,rombach2022sd,ramesh2022hierarchical} to progressively transform noise into coherent images through iterative denoising has established them as a powerful framework for generating high-quality, realistic images.
Despite their success, diffusion models are prone to certain flaws, with one persistent issue being the emergence of \textit{hallucinations} ---  generation of unrealistic or incorrect features that deviate from the training data distribution. 

Hallucinations can significantly degrade the quality of generated images, undermining the reliability of these models in real-world applications.
The increasing prevalence of AI-generated synthetic images online \citep{bender2023peacock} further compounds this issue, as \citet{aithal2024understanding} highlight that training future models on such data (some of which may include hallucinations), leads to performance degradation and can eventually collapse. 
Therefore, mitigating hallucinations is essential for preserving the fidelity of generated images and enhancing their practical utility.

Surprisingly, there has been little to no focused work on directly addressing hallucinations in diffusion models, especially in unconditional generation settings. 
The pioneering work of \citet{kim2024tackling} mitigates hallucinations through multiple local diffusion processes; however, it does not extend to unconditional generation. 
Additionally, the recent work of \citet{aithal2024understanding} explores hallucinations in unconditional models through the lens of mode interpolation and proposes a method for detecting them, however does not address their mitigation. 
Building on these foundations, we propose a novel approach to mitigating hallucinations in unconditional diffusion models, grounded in an analysis of the self-attention mechanism within the denoising/generative process of diffusion models.
While \citet{kim2024tackling} focus on tackling hallucinations that arise in image-conditional diffusion models due to out-of-domain conditioning (\eg, in tasks like super-resolution), the type of hallucinations we address is fundamentally different. 
Our work aims to mitigate hallucinations that occur {\it organically} in unconditional generative models. 
This scope of hallucinations aligns more closely with the phenomenon described by \citet{aithal2024understanding}, who provide explainability for these naturally occurring hallucinations through mode interpolation.

\begin{figure*}[t]
    \centering
    \includegraphics[width=\linewidth]{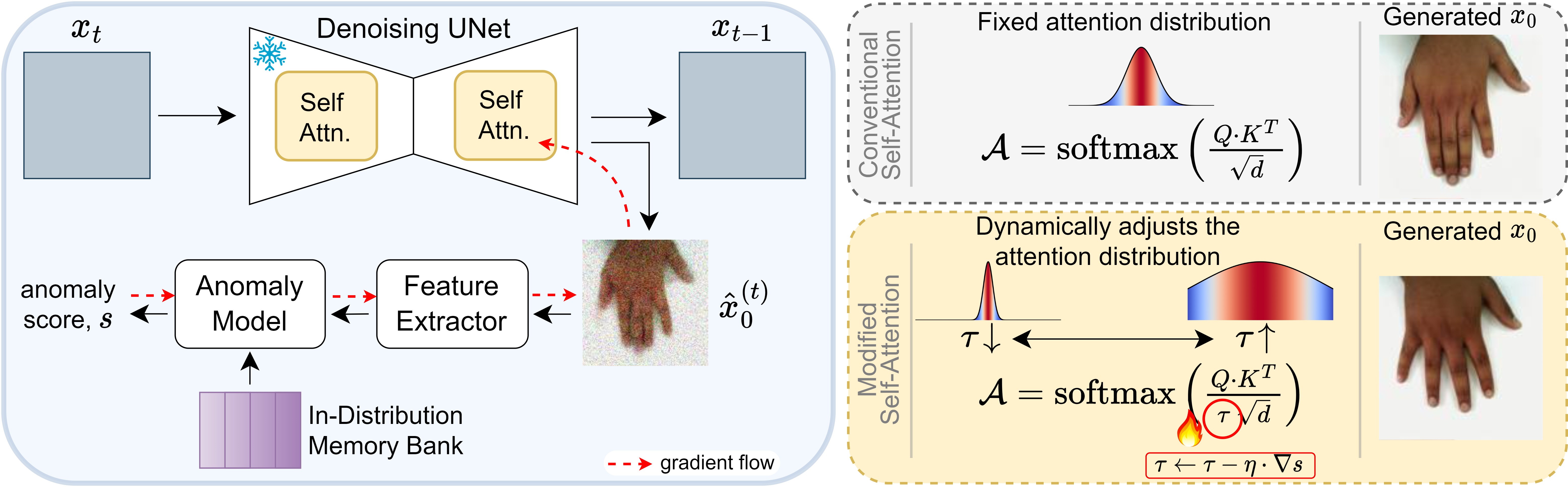}
    \caption{\textbf{Overview of our proposed pipeline for mitigating hallucinations in diffusion models with adaptive temperature scaling.} At each denoising timestep $t$, the noisy input $x_t$ is processed through a denoising UNet comprising self-attention layers. Conventional self-attention (\textit{top-right}) employs a fixed attention distribution, which may inadvertently amplify or suppress features, leading to hallucinations in the generated image $x_0$. In contrast, our approach (\textit{bottom-right}) introduces a temperature-scaled self-attention, where the temperature $\tau$, is dynamically adjusted based on an anomaly score $s$ from an anomaly detection model. This adaptive mechanism modulates the attention distribution, reducing noise-induced artifacts while preserving essential features, thereby enhancing the overall realism and fidelity of the generated images. }
    \label{fig:pipeline}
    \vspace{-5pt}
\end{figure*}

Our initial experiments led us to hypothesize that self-attention layers during early denoising steps play a critical role in amplifying or suppressing candidate noisy \textit{blobs} within an image.
These blobs, which represent potential image features, can become the seeds for hallucinations when improperly attended to during early denoising stages. 
In essence, we find that the way self-attention emphasizes or deemphasizes certain regions can lead to the manifestation of undesired or unrealistic content in the generated image.

To address this, we propose \textit{Adaptive Attention Modulation (AAM)}, a novel mitigation strategy centered around the introduction of a temperature parameter within the softmax operation of the self-attention layers.
This temperature parameter modulates the attention distribution, controlling the effective radius that each pixel can attend to. 
By carefully tuning this parameter, we can prevent unwarranted emphasis or suppression while ensuring that in-distribution content is preserved. 
This fine balance plays a crucial role in mitigating the emergence of hallucinations.

Recognizing that a single temperature parameter does not universally mitigate hallucinations, we employ an adaptive inference-time optimization to dynamically update the temperature as the denoising process progresses. 
While this approach substantially reduces hallucinations, we observed that certain blobs, particularly those manifesting in very early stages, can persist despite our temperature-based tuning. 
To address this, we introduce a complementary masked perturbation strategy, which involves selectively perturbing regions identified as potentially anomalous with random noise, effectively disrupting the propagation of these features through subsequent denoising steps.

To the best of our knowledge, this is the first work to address the issue of organically emerging hallucinations in diffusion models within an unconditional generation setting. 
In summary, our key contributions are as follows:
\begin{itemize}
    \item We investigate the role of self-attention in the manifestation of hallucinations during early denoising stages in diffusion models.
    \item We introduce a temperature scaling mechanism in self-attention layers as a simple yet effective approach to modulate the attention distribution, effectively controlling hallucinations.
    \item We propose a novel strategy to mitigate hallucinations by dynamically adjusting the temperature parameter, complemented with a masked perturbation technique. 
\end{itemize}
\section{Related Work}

\paragraph{Diffusion Models.}
Diffusion models, rooted in non-equilibrium thermodynamics theory, were introduced by \citet{sohl2015deep} and have since evolved into a robust framework for image generation.
The introduction of Denoising Diffusion Probabilistic Models (DDPMs) by \citet{ho2020ddpm} significantly advanced the field by demonstrating high-quality image synthesis through efficient reparameterization and objective functions. This sparked widespread interest and further development, including score-based diffusion by \citet{song2020score} that unified score-based generative modeling with diffusion processes, and improvements such as DDIM \citep{song2020ddim} and ADM \citep{nichol2021improved}, which reduced sampling steps for faster generation.

Conditional diffusion models have also gained traction, with guided-diffusion \citep{dhariwal2021guideddiffusion} enabling class-conditional synthesis and advancements like Imagen \citep{saharia2022imagen} and DALLE-2 \citep{ramesh2022dalle2} for text-to-image generation. 
Latent diffusion models introduced by \citet{rombach2022sd} further increased efficiency by operating in compressed latent spaces while maintaining quality.

Beyond image synthesis, diffusion models have been adapted for inpainting \citep{lugmayr2022repaint, corneanu2024latentpaint}, super-resolution \citep{saharia2022image, gao2023implicit}, and image translation \citep{saharia2022palette, wolleb2022swiss}. 
Furthermore, latent representations learned by diffusion models have found use in discriminative tasks, such as semantic segmentation, classification, and anomaly detection \citep{mukhopadhyay2023text, graikos2022diffusion, zimmermann2021score, wyatt2022anoddpm, pinaya2022fast}.

\vspace{-10pt}
\paragraph{Hallucinations.}  Diffusion models, despite their success, still suffer from failure modes, including hallucinations \citep{borji2023qualitative,narasimhaswamy2024handiffuser}, limited compositionality \citep{conwell2023comprehensive,gokhale2022benchmarking}, model memorization \citep{carlini2023extracting,somepalli2023diffusion}, and incoherent conditional generation \citep{liu2023discovering}. 
Hallucinations, in particular, refer to the generation of content that is not grounded in or consistent with the training data distribution, severely degrading the reliability and fidelity of generations. 
This flaw significantly hinders the practical applicability of diffusion models in real-world scenarios. 
\looseness=-1

Hallucinations have been extensively studied in large language models (LLMs) \citep{huang2023survey, tonmoy2024comprehensive, zhang2023siren} and large vision-language models (LVLMs) \citep{liu2024survey, guan2023hallusionbench, bai2024hallucination}, but remain relatively underexplored in image-focused diffusion models. 

\citet{kim2024tackling} tackled structural hallucinations in conditional generation by segmenting and processing out-of-distribution (OOD) regions separately. However, their approach is limited to image-conditional settings.
\citet{aithal2024understanding} studied hallucinations in unconditional diffusion models, attributing them to mode interpolation, where the model interpolates between data modes, leading to artifacts. 
They further observed that high variance in intermediate denoised outputs across successive timesteps signals emergence of hallucinations, implying a degree of self-awareness.
However, they did not propose a mitigation strategy.

Building on these insights, we hypothesize that self-attention layers in early denoising stages facilitate hallucinations by inappropriately amplifying or suppressing noisy features. 
To counter this, we introduce a dynamic temperature scaling to refine the attention distribution, focusing on relevant features and reducing artifacts.

\vspace{-10pt}
\paragraph{Temperature-Scaled Self-Attention.} Self-attention has become a transformative mechanism in machine learning, enabling models to assign varying importance to different parts of an input sequence, effectively capturing long-range dependencies and contextual relationships \citep{vaswani2017attention}.  
In vision tasks, self-attention enables the modeling of global interactions across an image, enhancing feature coherence. 
More recently, temperature scaling within self-attention mechanisms has emerged as a method to control the sharpness of the attention distribution, thereby influencing which correspondences are emphasized or suppressed.

Previous research has demonstrated the benefits of temperature-scaled self-attention in various contexts. 
\citet{zhang2021attention} used temperature scaling to smooth the attention distribution of teacher models in abstractive summarization, while \citet{lin2018learning} applied a self-adaptive approach to enhance language translation quality. 
In computer vision, \citet{zhou2023learning} incorporated temperature scaling to better utilize distant contextual information for image inpainting, and \citep{anonymous2025} employed temperature scaling to control the level of correspondence between the query and prompt image in visual in-context learning. 

Inspired by these successes, we introduce an adaptive temperature-scaled self-attention strategy within the denoising process of diffusion models. 
Our approach dynamically adjusts the temperature parameter based on an anomaly score, ensuring that attention is focused on relevant features while mitigating the amplification of noise. 
This technique offers a novel approach to mitigating hallucinations, effectively harnessing the strengths of self-attention while addressing its limitations.
\section{Method}

\subsection{Preliminaries}

Diffusion models \citep{ho2020ddpm} generate high-quality images by progressively transforming random noise into coherent images through a process of iterative denoising. 
These models rely on a denoising UNet that learns to reverse a forward noise addition process, with the goal of recovering the original image. 
The denoising UNet, parameterized by $\theta$, models an approximate posterior distribution:
\begin{equation}
    p_{\theta}(x_{t-1} | x_t ) =  \mathcal{N}\left ( x_{t-1}; \mu_\theta (x_t, t), \Sigma_\theta(x_t,t) \right ).
\end{equation}
At each timestep $t$, the UNet takes the noisy input $x_t$ and predicts the noise component $\epsilon_t$ that was added, obtaining the denoised output, $x_{t-1} = x_{t} - \epsilon_t$.
This denoising process is repeated for a finite number of steps, $T$, starting from random noise ($x_T \sim \mathcal{N}(\mu, \Sigma)$), progressively generating a realistic image, $x_0$. 

The denoising UNet incorporates multiple self-attention layers \citep{vaswani2017attention} at different resolutions, which are essential for capturing spatial dependencies across various regions of the image. 
This enables the model to produce locally and globally coherent features throughout the denoising process.

Within each self-attention layer, the intermediate features are projected into Query ($Q$), Key ($K$), and Value ($V$) vectors, which are then used to compute the feature updates, $\Delta\phi$, as follows:
\begin{equation}
    \Delta\phi = A \cdot V = \text{softmax}\left(\frac{Q\cdot K^T}{\sqrt{d}} \right ) \cdot V, \label{eq:feat_update_default}
\end{equation}
where, $A$ denotes the attention matrix and $d$ is the feature dimension of $Q$. 
The $i^{th}$ row of the attention matrix represents the weighted similarity between spatial location $i$ and all spatial locations in the image, including itself, indicating the relevance of each region of the image to $i$.

\subsection{Preliminary Observations}

\begin{figure}[t]
    \centering
    \includegraphics[width=0.65\linewidth]{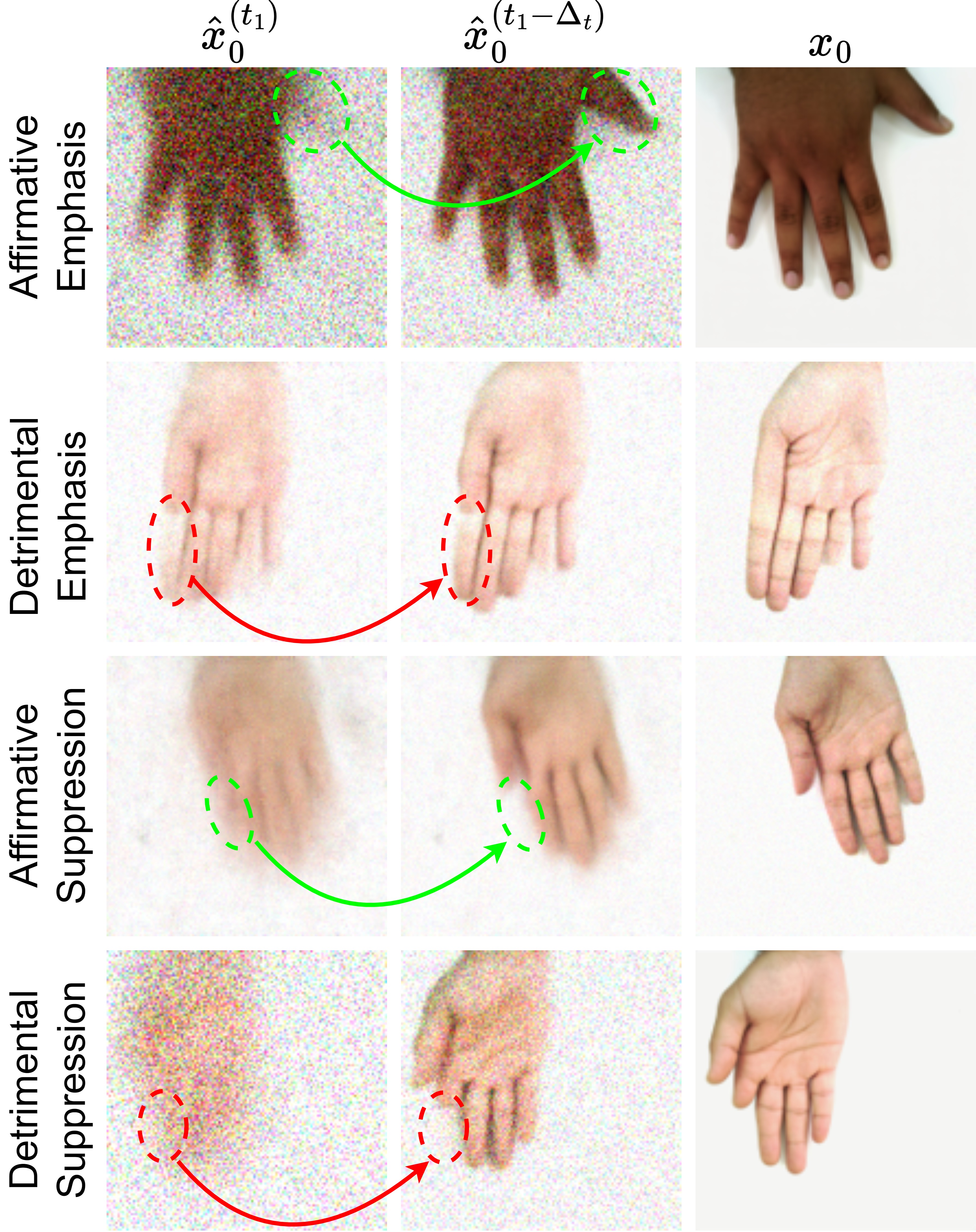}
    \caption{\textbf{Examples of affirmative and detrimental attention behaviors in diffusion models.}  The columns $\hat{x}_0^{(t_1)}$ and $\hat{x}_0^{(t_1-\Delta_t)}$ show the intermediate denoised predictions at an early denoising step $t_1$ and a subsequent step $t_1 - \Delta_t$. The $x_0$ column depicts the final denoised image. Rows 1 and 3 illustrate affirmative emphasis and suppression, respectively, resulting in realistic denoised images,  while rows 2 and 4 show detrimental emphasis and suppression, leading to hallucinations in the generated images. }
    \label{fig:posneg_emphsupp}
    \vspace{-5pt}
\end{figure}

Self-attention plays a crucial role, particularly in the early denoising steps of diffusion models, where the image is still dominated by noise.
During these stages, self-attention enables the model to identify and enhance relevant features while attenuating extraneous noise.
However, in certain cases, the attention mechanism may perform inversely (\ie amplifying irrelevant features or suppressing essential features), which may lead to hallucinations in the final generated image.
We illustrate examples of these two scenarios in \cref{fig:posneg_emphsupp}, where intermediate denoised predictions are shown. 
The intermediate prediction at timestep $t$ is obtained using the following formula:
\begin{equation}
    \hat{x}_0^{(t)} = \left ( x_t - \sqrt{1 - \bar{\alpha}_t} \epsilon_t \right ) / \sqrt{\bar{\alpha}_t},
\end{equation}
where, $\bar{\alpha}_t$ is the cumulative product of the noise schedule up to timestep $t$ \citep{ho2020ddpm}. 
In \cref{fig:posneg_emphsupp}, rows 1 and 3 depict affirmative attention behaviors, where relevant features are correctly emphasized or irrelevant ones are suppressed, resulting in realistic images.
In contrast, rows 2 and 4 illustrate detrimental behaviors, where incorrect emphasis or suppression of features contributes to hallucinations\footnote{We use examples from hand generation, as hallucinatory artifacts like missing/additional/malformed fingers are particularly easy to identify.}.

To investigate and potentially control these unintended behaviors, we introduce a temperature parameter, $\tau$, to modulate the sharpness of the attention distribution in the softmax normalization. 
By adjusting $\tau$, we can control the spread of attention across spatial locations, potentially reducing unwanted emphasis or suppression. 
Formally, incorporating $\tau$, modifies the feature update equation, \cref{eq:feat_update_default} as follows:
\begin{equation}
    % \vspace{-5pt}
    \Delta \phi = \text{softmax} \left ( \frac{Q \cdot K^T}{\tau \cdot \sqrt{d}}\right) \cdot V.
\end{equation}

\vspace{-10pt}
\paragraph{Effect of Temperature Scaling on Attention Distribution.}
To understand how $\tau$ affects the attention distribution, we start with the softmax function, which transforms the similarity scores between Query ($Q$) and Key ($K$) vectors into a probability distribution, thereby computing the attention weights.
Expanding the softmax function, the attention distribution is given by,
\begin{equation}
    A_{i,j} = \frac{\text{exp} \left ( \frac{Q_i \cdot K_j}{\tau \cdot \sqrt{d}} \right )}{\Sigma_m \text{exp} \left ( \frac{Q_i \cdot K_m}{\tau \cdot \sqrt{d}} \right )},
\end{equation}
where, $A_{i,j}$ represents the attention weight between spatial locations $i$ and $j$. 
The formulation reduces to the original attention distribution when $\tau=1.0$, effectively removing the temperature scaling. 

As $\tau$ decreases, the factor $1/\tau$ increases, making the exponentials more sensitive to high similarity scores. This results in a sharper distribution, where a few spatial locations receive most of the attention, and the distribution approaches a point mass as $\tau \rightarrow 0$.
Conversely, as $\tau$ increases, the exponents become less sensitive to differences in similarity scores, resulting in a more uniform attention distribution, where the model attends to a broader range of spatial locations.
This temperature scaling mechanism allows us to modulate the radius of focus in self-attention, which can help control the amplification of relevant or irrelevant features in early denoising stages. 

\begin{figure}[t]
    \centering
    \includegraphics[width=0.8\linewidth]{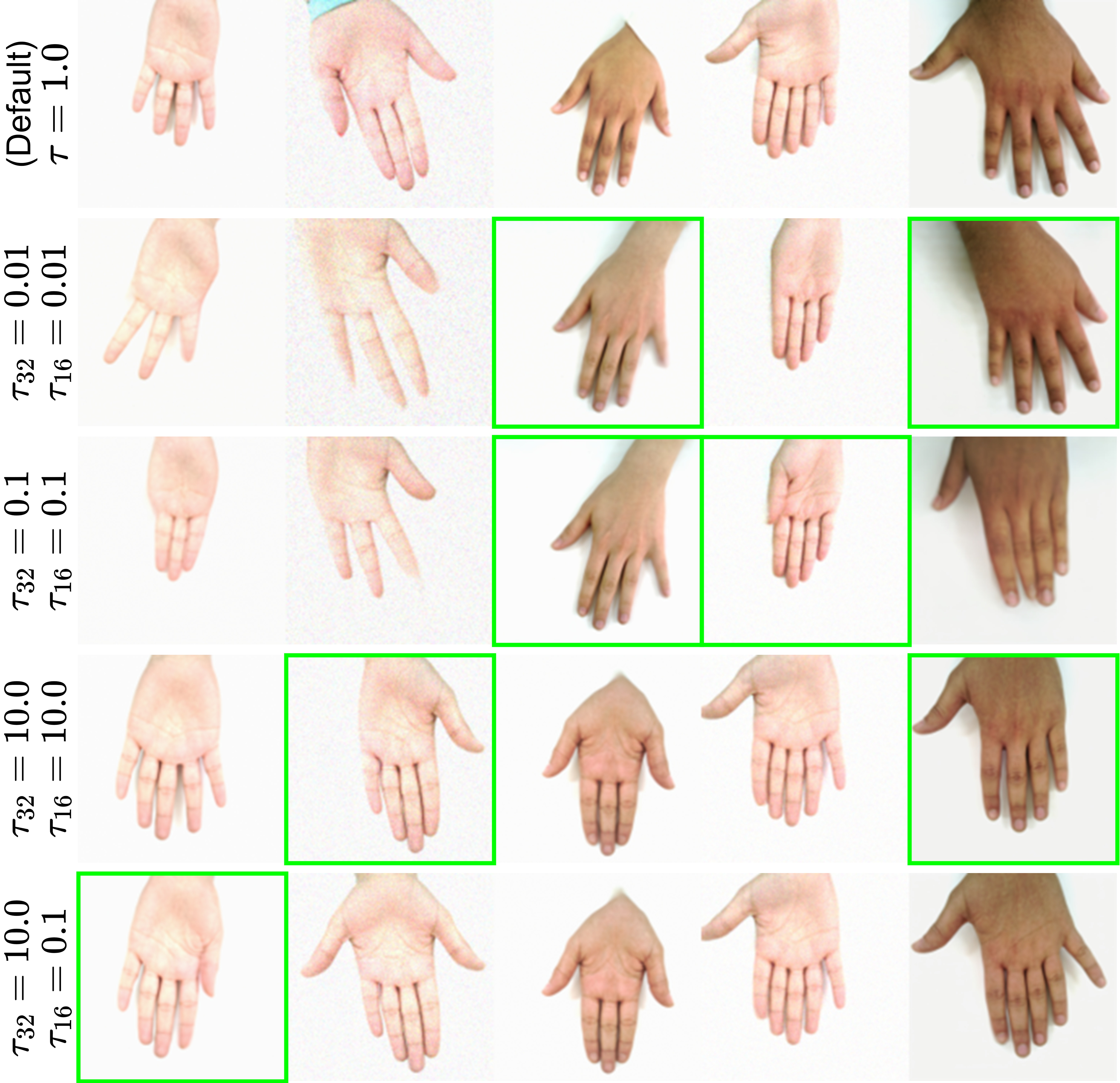}
    \caption{\textbf{Example generations with temperature scaled attention.} For each column, we start with the same initial noise ($x_T$), which results in hallucinations under the default setting with no temperature scaling (\ie $\tau=1.0$). Each row depicts the generated samples for different combinations of $\tau_{32}$ and $\tau_{16}$, corresponding to the resolutions at which the self-attention layers operate --- $32\times32$ and $16\times 16$, respectively. Hallucination-free generations are indicated with green borders. This figure demonstrates that adjusting $\tau$ can mitigate hallucinations, although no single temperature value provides a universal solution across all scenarios. }
    \label{fig:temp_test}
    \vspace{-8pt}
\end{figure}

\paragraph{Empirical Observations on Temperature Scaling.}
Initially, we experimented with different values of $\tau$ to observe its impact on mitigating hallucinations using the same set of initial noise ($x_T$) across trials. 
For reproducibility, we used DDIM sampling \citep{song2020ddim}, which enables deterministic denoising. 
Our observations, shown in \cref{fig:temp_test}, indicate that by manually adjusting $\tau$, we could mitigate hallucinations and produce in-distribution images from initial conditions that previously led to hallucinations. 

However, as observed in \cref{fig:temp_test}, these experiments reveal that there is no single, fixed value of $\tau$ that {\it consistently} prevents hallucinations across different scenarios. 
While a specific $\tau$ may improve the outcome for one sample, it can exacerbate hallucinatory artifacts in another. 
This limitation indicates that a static temperature value cannot robustly address hallucinations universally. 

\subsection{Adaptive Attention Modulation (AAM)}

Our preliminary experiments suggest that a dynamic, context-aware tuning of the temperature parameter $\tau$ is essential for robust hallucination mitigation, as manual adjustment is impractical due to its time-consuming and labor-intensive nature. 
Therefore, we propose an adaptive temperature tuning mechanism that dynamically adjusts $\tau$ during denoising to modulate the attention distribution, ensuring effective hallucination mitigation across diverse scenarios. 

\vspace{-10pt}
\paragraph{Inference-Time Optimization for Temperature Tuning.}
We implement adaptive temperature tuning through an inference-time optimization strategy, a technique successfully employed in prior work \citep{poole2022dreamfusion, burgert2022peekaboo, burgert2024diffusion, lin2023magic3d}.
By optimizing the temperature parameter $\tau$ on-the-fly, our method tailors the attention distribution's sharpness to the specific context of each denoising step.

To ensure $\tau$ stays within a practical and stable range, we optimize an intermediate variable, $\hat{\tau}$, rather than directly optimizing $\tau$. 
Specifically, we define $\tau = 10^{\gamma \cdot \tanh{(\hat{\tau}})}$, allowing $\tau$ to vary smoothly within a controlled range (\ie $\tau \in [10^{-\gamma}, 10^{\gamma}]$), preventing extreme values that could destabilize the denoising process. 
We employ an exponential transformation because we observed that meaningful changes in $\tau$ often occur across different orders of magnitude, allowing for effective scaling at varying levels. 
This formulation also permits larger adjustments when needed, particularly in the early stages of optimization, while ensuring that $\tau$ remains positive, thereby avoiding potential issues associated with negative values.

An anomaly score derived from the PatchCore anomaly detection model \citep{roth2022patchcore}, which is a training-free model,  guides the optimization.
PatchCore constructs a memory bank from features of in-distribution training samples, which is then used to detect out-of-distribution (OOD) anomalies in generated images. 

To tailor PatchCore to our setup, we introduce a few modifications. 
To better match the early denoising stages, where noise is dominant, we augment the training data of the diffusion model with noisy samples generated using the same noise schedule as when training the diffusion model. 
This prevents the anomaly detection model from falsely identifying these noise characteristics as OOD. 
Further, instead of extracting features from an ImageNet pre-trained WideResNet-50 \citep{zagoruyko2016wide}, we extract features from the diffusion model's denoising UNet. 
These features are more representative of the intermediate outputs, as they are specifically tuned to extract meaningful structures from noisy inputs.
Following \citet{roth2022patchcore}, we use mid-level features (specifically from layers 7 and 11 of the UNet encoder), as they provide a balance, being neither too generic (as shallow features might be) nor overly biased towards denoising (as deeper features could be). 

Our method dynamically tunes $\tau$ over specific denoising timesteps to adapt to the evolving structure of the image as it progresses through the generation process:
\begin{enumerate}
    \item \textbf{Initial Denoising Stage ($T \ge t > T_1$)}:
    For the earliest timesteps, the model denoises using the default temperature setting (\ie $\tau=1.0$). 
    This phase allows the model to establish foundational structures from initial random noise, ensuring that the anomaly model has structured features to analyze rather than pure randomness.

    \item \textbf{Adaptive Tuning Stage  ($T_1 \ge t > T_2$)}: 
    In this stage, high-level structure and coarse features are defined, as observed in prior work \citep{choi2022perception, park2024explaining}.
    As hallucinations are most likely to emerge during this stage, we optimize $\tau$ during this range of timesteps as the coarse structure forms.
    The optimization details are as follows: 
    \begin{itemize}
        \item The optimization is performed over $N$ iterations, with early stopping based on the gradient norm to improve computational efficiency.
        \item At each timestep $t$, $\tau$ is initialized to the optimal value from the previous step. Further to prevent being stagnant at a local minimum, $\tau$ is periodically re-initialized to the default value.
        \item The objective is to minimize the anomaly score predicted by PatchCore, ensuring that features inconsistent with in-distribution characteristics are minimized during denoising.   
    \end{itemize}

    \item \textbf{Final Denoising Stage ($T_2 \ge t > 0$)}: 
    Once the high-level structure is established, we revert to the default temperature setting for the remaining timesteps.
    Since these later steps primarily focus on refining low-level/finer details (\eg textures), hallucinations are less likely to emerge, making adaptive tuning unnecessary during this phase.
\end{enumerate}

\vspace{-10pt}
\paragraph{Masked Perturbation for Early Hallucination Suppression.}
While adaptive temperature tuning addresses most hallucinations, certain hallucinatory artifacts manifesting during the very early denoising stages (\ie during $t \in [T,T_1]$) can persist. 
Since intermediate outputs at this stage are predominantly noise, these early hallucinations cannot be effectively detected by the anomaly detection model. 
To handle these cases, we introduce a masked perturbation strategy that selectively disrupts emerging hallucinations in localized regions. 

At specific denoising timesteps $t \in L$, we compute an anomaly heatmap $h$ from PatchCore, generating a binary mask $M$ based on a threshold $\beta$. 
This binary mask highlights anomalous regions, which are selectively perturbed as follows: 
\begin{equation}
    x_{t-1} \leftarrow M \cdot \zeta + (1-M) \cdot x_{t-1}, \quad \zeta \sim \mathcal{N}(\mu, \Sigma).
\end{equation}
This operation introduces noise selectively into the anomalous regions, preventing early-stage hallucinations from persisting in subsequent steps.

The complete algorithm for adaptive temperature tuning with masked perturbation is outlined in \cref{algo:adaptive_tuning}.

\begin{algorithm} 
\caption{Adaptive Attention Modulation}
\label{algo:adaptive_tuning}
\SetKwInOut{KwInit}{Initialize}

\KwIn{Diffusion model $f_{\theta}(\cdot)$, PatchCore model $g(\cdot)$, num. of timesteps $T$, thresholds $\lambda$, $\beta$, $\delta$, num. of optimization steps $N$, learning rate $\eta$, perturbation set $L$ }

\KwOut{Generated image, $x_0$}

\KwInit {$x_T \sim \mathcal{N}(\mu, \Sigma) $} 

\For{$t = T$ \textbf{down to} $1$}{
    \If{$T_1 \ge t > T_2$ }{ \tcp{\footnotesize Adaptive Tuning Stage}
        \If{$(T_1-t) \mod \lambda = 0$}{
            \tcp{\footnotesize Periodical Re-initialization}
            $\hat{\tau} \gets 0$ \; 
        }
        \For{$n = 1$ \textbf{to} $N$}{
            \tcp{\footnotesize Optimizing $\hat{\tau}$}
            $\epsilon_t = f_\theta(x_t, \tau= 10^{\gamma \cdot \tanh(\hat{\tau})})$\;
            $\hat{x}_0^{(t)} = \left ( x_t - \sqrt{1 - \bar{\alpha}_t} \epsilon_t \right ) / \sqrt{\bar{\alpha}_t}$ \;
            $s,h = g(\hat{x}_0^{(t)})$ \;
            $\hat{\tau} \gets \hat{\tau} - \eta \cdot \nabla_{\hat{\tau}} s$ \;
            \If{$|\nabla_{\hat{\tau}} s| <  \delta$}{
                \textbf{break}
            }
        }

        $\epsilon_t = f_\theta(x_t, \tau = 10^{\gamma \cdot \tanh(\hat{\tau})})$\;
        $x_{t-1} = x_{t} - \epsilon_{t}$ \;
        $\hat{x}_0^{(t)} = \left ( x_t - \sqrt{1 - \bar{\alpha}_t} \epsilon_t \right ) / \sqrt{\bar{\alpha}_t}$ \;
        \If{$t \in L$}{
            \tcp{\footnotesize Masked Perturbation}
            $s,h = g(\hat{x}_0^{(t)})$ \;            
            $M = h > \beta$ \;
            $x_{t-1} \gets M \cdot \zeta + (1 - M) \cdot x_{t-1}$ \;
        }
    }

    \Else{
        \tcp{\footnotesize Default Temperature Setting}
        $\epsilon_t = f_\theta(x_t, \tau = 1.0)$\;
        $x_{t-1} = x_{t} - \epsilon_{t}$ \;
    }
}
\end{algorithm}
\section{Experiments and Results}

\begin{figure*}[t]
    \centering
    \includegraphics[width=\linewidth]{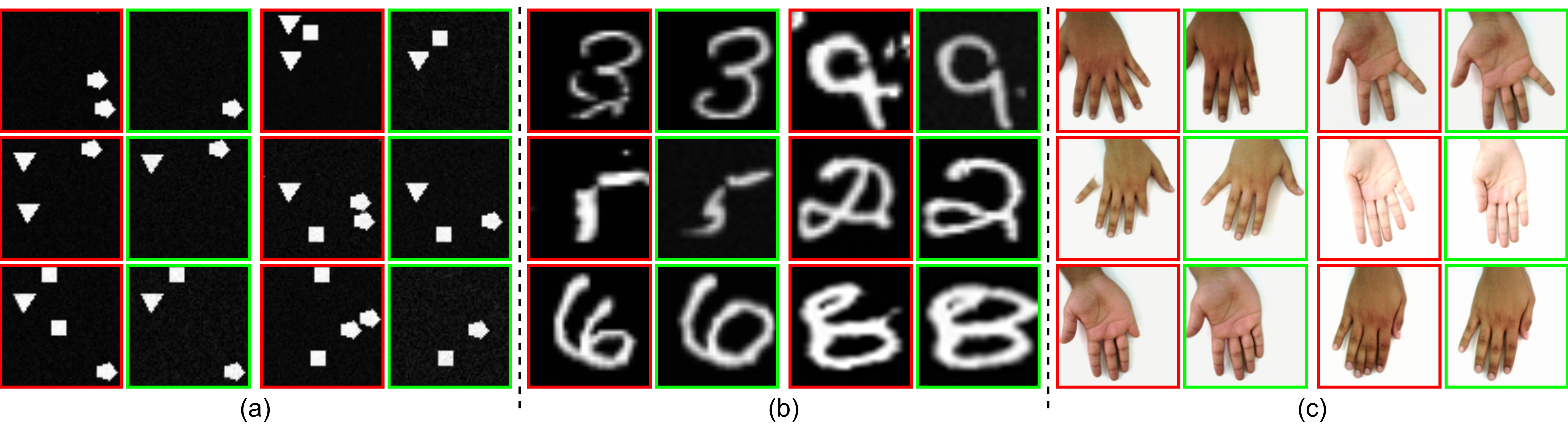}
    \caption{\textbf{Effectiveness of hallucination mitigation across datasets.} Visual comparison of generated samples with (green border) and without (red border) our proposed method. From left to right: (a) Simple Shapes dataset \citep{aithal2024understanding}, where our method prevents the generation of extra instances, adhering to the single-instance-per-region distribution; (b) MNIST dataset \citep{lecun1998gradient}, where hallucinated artifacts that distort the digits are reduced; (c) Hands dataset \citep{afifi201911k}, where hallucinations such as missing, extra, or malformed fingers are corrected, producing anatomically accurate hands. Our approach consistently enhances realism and fidelity in generated images across different datasets.}
    \label{fig:results}
    \vspace{-5pt}
\end{figure*}

\subsection{Experimental Details}

We evaluate our proposed approach on three datasets: Simple Shapes \citep{aithal2024understanding}, MNIST \citep{lecun1998gradient}, and Hands \citep{afifi201911k}, each previously used in \citet{aithal2024understanding} to interpret the causes of hallucinations in diffusion models. 
\begin{itemize}
    \item \textbf{Simple Shapes:} This synthetic dataset comprises 5,000 black-and-white images, where each image is divided into three equal vertical regions designated for a triangle, square, and pentagon, respectively. 
    Each shape has a 0.5 probability of appearing in its assigned column. 
    Following \citet{aithal2024understanding}, we train an unconditional DDPM \citep{sehwag2023minimaldiffusion} on this dataset. 
    For this model, we define hallucinations as instances where multiple shapes appear within a single column region (\eg \cref{fig:results} (a)). 

    \item \textbf{MNIST:} This dataset, widely used for handwritten digit recognition, consists of 60,000 grayscale images. 
    We train a class-conditional DDPM trained with classifier-free guidance \citep{ho2022cfg}. 
    Hallucinations observed in this model manifest as artifacts that distort the digits, potentially impacting recognition (\eg \cref{fig:results} (b)). 

    \item \textbf{Hands:} We randomly sample 5,000 images from the Hands dataset, which consists of high-quality images of human hands against uniform backgrounds.
    For this dataset, we train an ADM model \citep{nichol2021improved}.
    Hallucinations in this dataset are visually prominent, often manifesting as deformations or unrealistic anomalies, such as missing, additional, or malformed fingers (\eg \cref{fig:results} (c)).
\end{itemize}

For each dataset, we generate 1,000 samples using the same initial noise vectors and employ DDIM sampling with 250 inference steps, ensuring consistency and reproducibility across all experiments. 
We evaluate these generated images of all these datasets using FID \citep{heusel2017gans}. 
Further, since hallucinations are easily identifiable in Shapes and Hands datasets we report the percentage of hallucinated images directly. 
However, in MNIST due to the subjective difficulty in identifying hallucinated artifacts in handwritten digits, we instead report the classification accuracy.  

\subsection{Implementation Details}

Our implementation uses the following empirically determined hyperparameters unless otherwise specified. 
All models are trained with $T=1000$ timesteps. 
We focus on optimizing the temperature parameter, $\hat{\tau}$, between timesteps $T_1 = 0.92T$ and $T_2=0.6T$, corresponding to the interval where \citet{aithal2024understanding} detect hallucinations via high-variance intermediate outputs.
A scaling factor $\gamma=2$ is used to optimize $\hat{\tau}$ in the range of $[-2 , 2]$, effectively mapping the temperature parameter to the interval $[0.01, 100]$.
We use the Adam optimizer \citep{kingma2014adam} with a learning rate $\eta=0.01$ and run the optimization for $N=10$ iterations, applying early stopping when the gradient norm drops below $\delta=0.001$.

We periodically reinitialize $\hat{\tau}$ at intervals of $\lambda = 0.04T$, and masked perturbations are introduced just before each of the first three reinitializations: \ie $L = \{ T_1 - \lambda k + 1 ; k \in \{0,1,2\}\}$. We avoid further perturbations beyond this point, as they impede the model’s ability to recover meaningful features from the perturbation.  
The threshold for masking is set at $\beta = \mu_s + 1.5 \sigma_s$, where $\mu_s$ and $\sigma_s$ denote the mean and standard deviation of anomaly scores derived from the training samples.  

\subsection{Results and Discussion}

As outlined in the introduction, our work is the first, to the best of our knowledge, to propose a method for mitigating hallucinations that occur organically in unconditional generative models. 
Consequently, we evaluate the effectiveness of our approach by comparing it to a standard diffusion model without temperature scaling, which serves as our baseline. 
These results are summarized in \cref{tab:quant_results}, and visual examples are depicted in \cref{fig:results}.
Our method demonstrates substantial improvements in FID scores across all datasets: 12.1\% in Shapes, 25.6\% in MNIST, and 20.8\% in Hands. 
Additionally, the percentage of hallucinated images decreases by 6.7\% in the Shapes dataset and 12.9\% in Hands dataset, while classification accuracy  in MNIST improves by 2.3\%.

\begin{table}[tbp]
    \centering
    \resizebox{\linewidth}{!}{
    \begin{tabular}{c|cc|cc|cc}
        \toprule
        \multirow{2}{*}{Method} & \multicolumn{2}{c|}{Shapes} & \multicolumn{2}{c|}{MNIST} & \multicolumn{2}{c}{Hands} \\
        & FID$\downarrow$ & Hal.\% $\downarrow$& FID$\downarrow$ & Acc.$\uparrow$ & FID$\downarrow$ & Hal.\% $\downarrow$\\
        \midrule
        Default & 178.4 & 7.8 & 20.3 & 92.0 & 129.1 & 22.1 \\
        Ours & 156.9 & 1.1 & 15.1 & 94.3 & 102.3 & 9.2\\
        \bottomrule
    \end{tabular}
    }
    \caption{\textbf{Quantitative evaluation of our proposed method compared to the default diffusion model without temperature scaling.} Our approach demonstrates improvements across all metrics, including enhanced FID scores for all datasets, higher classification accuracy for MNIST, and a reduced percentage of hallucinated images for both Shapes and Hands datasets.}
    \label{tab:quant_results}
    \vspace{-10pt}
\end{table}

\begin{table*}[tbp]
    \centering
    \begin{minipage}{0.3\textwidth}
        \centering
        \resizebox{\linewidth}{!}{%
            \begin{tabular}{l|c}
                \toprule
                Method & FID $\downarrow$ \\
                \midrule
                Default ($\tau=1.0$) & 129.1 \\
                \midrule
                Fixed $\tau$: & \\
                \quad $\tau = 0.01$ & 127.6 \\
                \quad $\tau = 0.1$ & 127.3 \\
                \quad $\tau = 10.0$ & 128.5 \\
                \midrule
                Adaptive $\tau$ & 115.2 \\
                \quad + Periodic re-initialization & 109.0 \\
                \quad + Masked perturbation & 102.3 \\
                \bottomrule
            \end{tabular}
        }
        \caption{Effect of temperature scaling configurations and sub-components in the proposed algorithm. }
        \label{tab:temp_param}
    \end{minipage}
    \hfill
    \begin{minipage}{0.3\textwidth}
        \centering
        \resizebox{\linewidth}{!}{%
            \begin{tabular}{ccc|c}
                \toprule 
                \multicolumn{3}{c|}{Resolution} & \multirow{2}{*}{FID $\downarrow$} \\
                $8\times 8$ & $16\times 16$ & $32 \times 32$ & \\
                \midrule
                \checkmark & - & - & 127.6\\
                - & \checkmark & - & 115.8\\
                - & - & \checkmark & 110.9\\
                \checkmark & \checkmark & - & 114.3\\
                \checkmark & - & \checkmark & 109.1\\
                - & \checkmark & \checkmark & 103.4\\
                \checkmark & \checkmark & \checkmark & 102.3\\
                \bottomrule
            \end{tabular}
        }
        \caption{Impact of modifying different attention resolutions with adaptive temperature scaling.}
        \label{tab:attn_res}
    \end{minipage}
    \hfill
    \begin{minipage}{0.34\textwidth}
        \centering
        \resizebox{\linewidth}{!}{%
            \begin{tabular}{l|c|c}
                \toprule
                Feature Backbone & Noised Aug. & FID $\downarrow$ \\
                \midrule
                \multirow{2}{*}{ResNet (ImageNet)} & \texttimes & 116.3\\
                                                   & \checkmark & 115.7\\
                \midrule
                \multirow{2}{*}{Diff. UNet (Hands)} & \texttimes & 105.5 \\
                                                   & \checkmark & 102.3\\
                \bottomrule
            \end{tabular}
        }
        \caption{Effectiveness of using the diffusion model’s denoising UNet for feature extraction compared to a ResNet backbone and the impact of noise augmentation for in-distribution memory bank creation.}
        \label{tab:memory_bank_ablation}
    \end{minipage}
    \vspace{-5pt}
\end{table*}

\begin{figure}[tbp]
    \centering
    \includegraphics[width=0.65\linewidth]{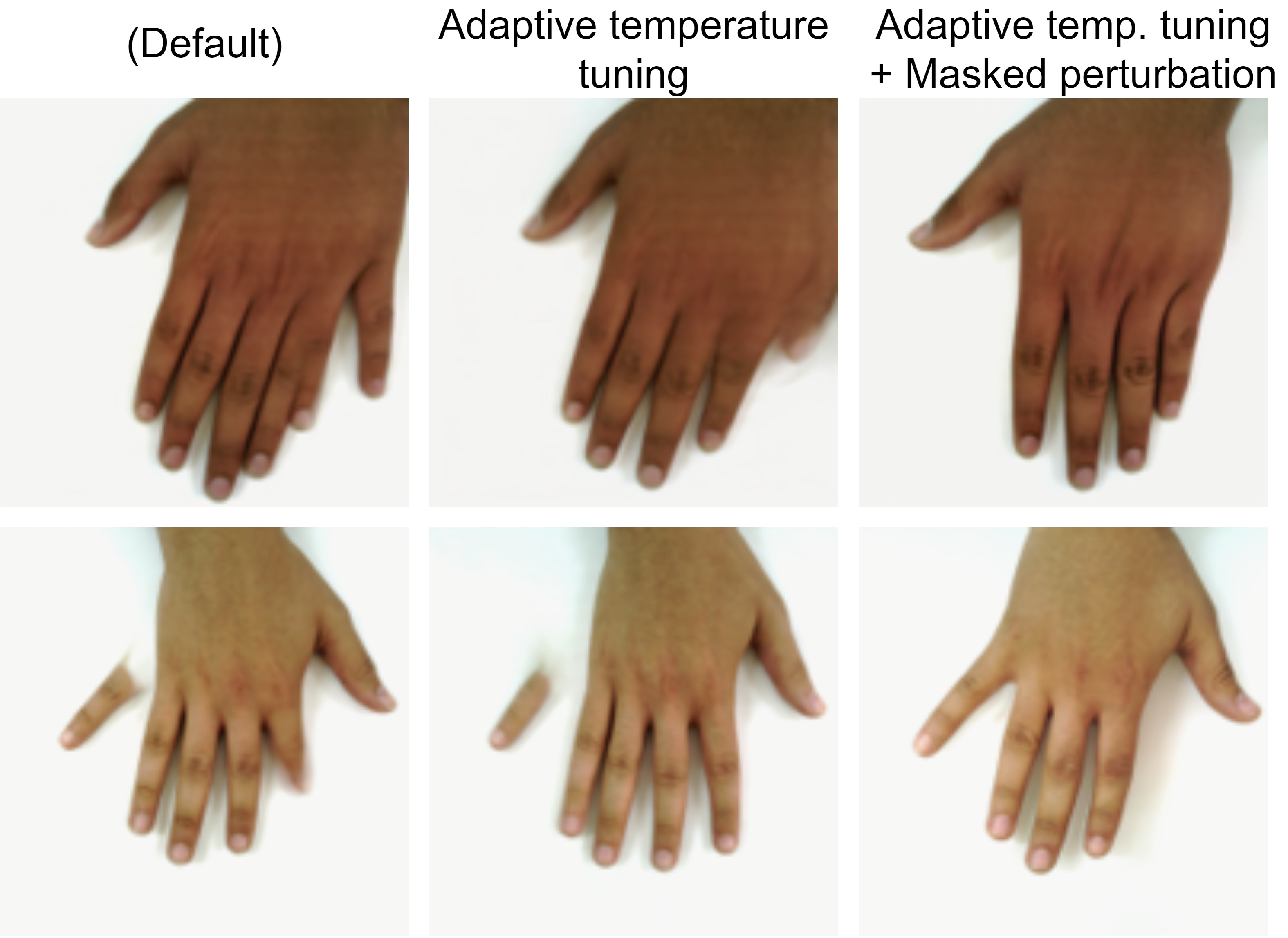}
    \caption{\textbf{Effect of adaptive temperature tuning and masked perturbation.} The default diffusion model (left) commonly generates artifacts like missing or extra fingers. Adaptive temperature tuning (middle) reduces such artifacts to a certain extent, while combining it with masked perturbation (right) effectively mitigates hallucinations, producing anatomically accurate hands.}
    \label{fig:abl_tempnoise}
    \vspace{-10pt}
\end{figure}

To gain more insights through ablations, which are discussed below, we concentrate on the Hands dataset, which better reflects real-world data.
In \cref{tab:temp_param} we illustrate the performance gains achieved by progressively incorporating each sub-component of our method, beginning with the baseline (a standard diffusion model equivalent to $\tau=1.0$).
First, we analyze the impact of using fixed temperature values ($\tau \in \{0.01, 0.1, 10.0\}$), which yield modest improvements over the baseline but show relatively consistent results.
This suggests that static temperature values are inadequate to handle varying noise conditions throughout the denoising process and diverse hallucination scenarios, as also reflected in the qualitative examples illustrated in \cref{fig:temp_test}.
Introducing a dynamically adjusted temperature parameter results in a notable reduction in FID scores, underscoring the importance of adaptive temperature control.
Additional performance gains are observed with periodic re-initialization, likely because it helps the process avoid getting stuck in sub-optimal states.
Finally, our masked perturbation strategy mitigates hallucinations through targeted perturbation of anomalous regions early in the denoising process, resulting in additional improvements in FID. 
In \cref{fig:abl_tempnoise}, we depict visual examples where adaptive temperature scaling alone is insufficient. 
In such cases, while adaptive temperature tuning partially reduces hallucinated features, combining it with masked perturbation effectively alleviates these artifacts, resulting in more realistic hand images. 

Additionally, we investigate the impact of modifying different combinations of attention resolutions ($8\times8$, $16\times16$, and $32\times32$) with adaptive temperature scaling in \cref{tab:attn_res}. 
Results indicate that modifying all resolutions yields the best performance; however, higher resolutions play a more substantial role in hallucination mitigation. 
This is intuitive, as lower resolutions primarily capture global features and their dependencies, while higher resolutions focus on finer, localized details essential for relatively accurate structure and feature delineation. 
By adjusting the attention distribution at higher resolutions, the model can better suppress hallucinatory artifacts that may arise from inaccurately emphasized local features.

We also explore the effectiveness of using the denoising UNet as the feature extraction backbone, compared to a ResNet as used in prior work \citep{roth2022patchcore,kim2024tackling}. 
In this context, we evaluate FID scores for feature extraction using an ImageNet-pretrained ResNet (specifically WideResNet-50 \citep{zagoruyko2016wide}) versus the diffusion model's encoder, with and without noise-augmented images when constructing the in-distribution memory bank.
The results which are tabulated in \cref{tab:memory_bank_ablation}, indicate that the denoising UNet consistently outperforms the ResNet backbone, likely because the UNet is trained to process noisy inputs, making it more effective at extracting meaningful features in noisy images. 
Additionally, using the denoising UNet itself to extract features eliminates reliance on external models.
Furthermore, noise augmentation in the memory bank creation process enhances performance, as it better simulates intermediate denoising outputs and improves memory bank robustness.

While our method successfully mitigates hallucinations, a limitation is the increased inference time due to the dynamic optimization of the temperature parameter, a common issue among inference-time optimization methods. 
A potential solution could involve incorporating a learnable temperature parameter directly into the training process, enabling the diffusion model to adaptively scale self-attention without additional inference-time adjustments. 
This approach could significantly reduce inference time while preserving performance gains.
Furthermore, the lack of a structured, standardized method to systematically generate and detect hallucinations constrains our experiments to a limited set of data distributions. 
Future work on structured frameworks for hallucination generation and detection could broaden the evaluation scope and advance mitigation techniques across varied datasets and diffusion models.

\section{Conclusion}

In this work, we propose a novel approach to mitigate hallucinations in diffusion models, specifically addressing artifacts that occur organically in unconditional generative settings.
Preliminary experiments reveal that self-attention layers in early denoising stages may inadvertently amplify or suppress noisy features, potentially leading to hallucinations.
Leveraging this insight, we propose a novel adaptive temperature scaling strategy, augmented by masked perturbations, to dynamically modulate the attention distribution, effectively reducing hallucinations. 
Experiments across datasets demonstrate significant improvements in FID scores and hallucination reduction, enhancing the fidelity and reliability of generated images.

{
    \small
    \bibliographystyle{ieeenat_fullname}
    \bibliography{main}
}

\end{document}